# JONES-19: A Cultural Image Dataset Based on *The Grammar of Ornament*

Linh Pham

Harvard University, linhpham@mde.harvard.edu

David Alvarez-Melis

Harvard University, dam@seas.harvard.edu

Stuart Shieber

Harvard University, shieber@seas.harvard.edu

Alexandros Haridis*

Harvard University, haridis@seas.harvard.edu

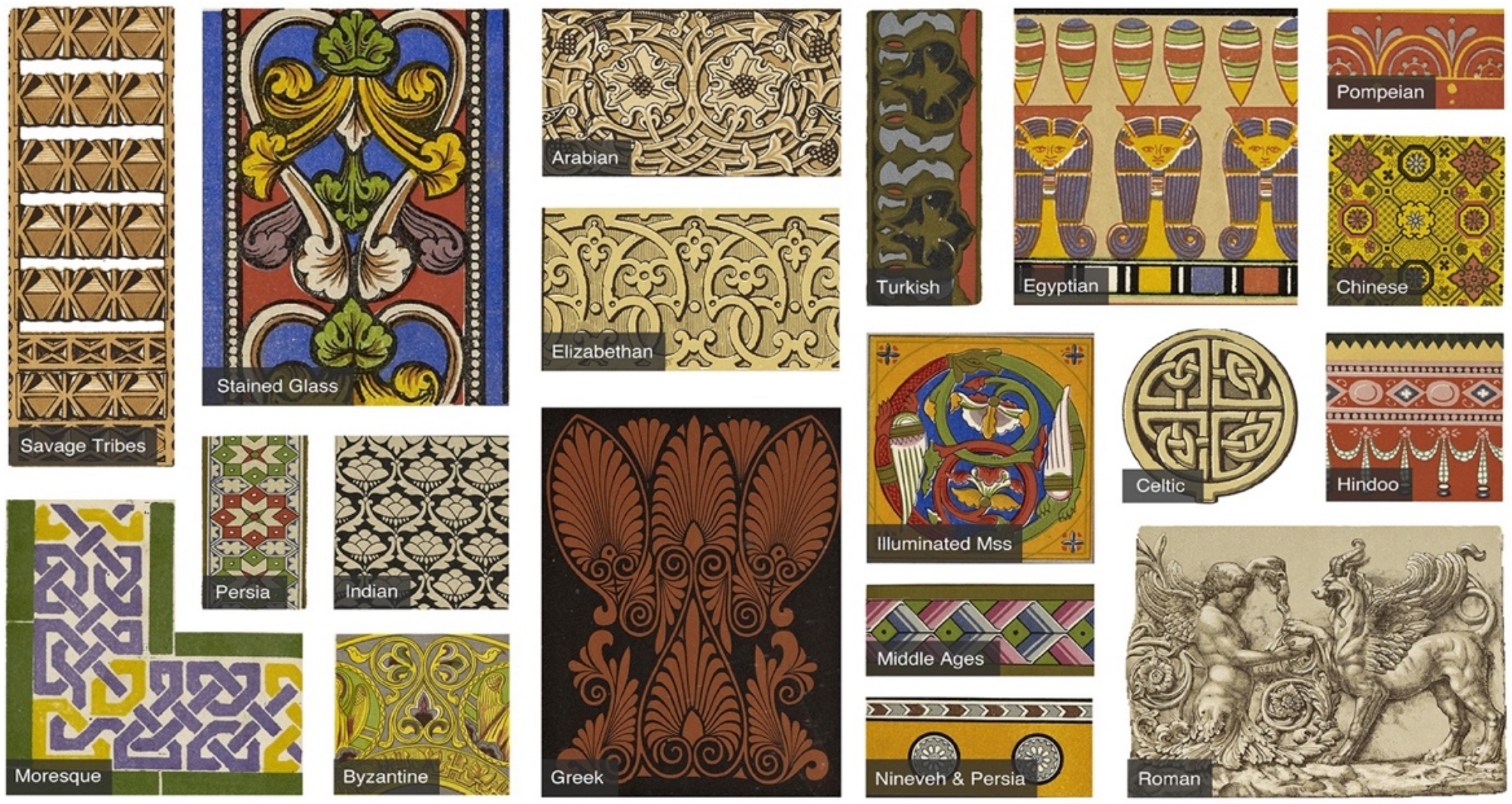

**Figure 1: The JONES-19 dataset consists of ornament designs documented in *The Grammar of Ornament* (London, 1857).**

## Abstract

We introduce JONES-19, a high-quality image dataset documenting 1,901 ornament designs belonging to nineteen human cultures. The images and their annotations are based on an open access archive of *The Grammar of Ornament* (London, 1856), by Owen Jones. The dataset poses numerous challenges as a benchmark for computer vision classification tasks and for research at the intersection of machine learning and art and design: a small sample size, image samples of human-designed artifacts rather than common objects in-context or natural scenes, imbalanced class distribution, and image distinctions based on fine details involving line patterns, reliefs, and colors. As a design-inspired dataset, JONES-19 can serve as a benchmark for various research fields, such as visual recognition, data-efficient learning, art-historical research, architectural style analysis, and cultural heritage. This paper describes the curation of the JONES-19 dataset and reports a baseline classification benchmark that evaluates the suitability of the dataset for training classifiers and exposes insights into inter-class relationships—particularly cultural similarities—by examining patterns in misclassification errors. The dataset and its accompanying documentation are available at: https://huggingface.co/datasets/harvardseas-cultural-ornaments/JONES-19.

* Corresponding author. Current email address: charidis@mit.edu

## 1 INTRODUCTION

The limited use and success of machine learning (ML) methods on design or cultural collections can be attributed to issues related to access and quality of available data sources. The history of architecture and art, however, contains numerous well-documented instances of design artifacts, ranging from ornaments and patterns to building and furniture designs. The digitized or "virtual" manifestations of these documented artifacts, in conjunction with the newest advances in data-driven ML technology, have the potential to introduce new and important frontiers for discriminative or generative modeling research in ML as well as for the analysis or generation of design and cultural collections. This paper introduces JONES-19, a new high-quality image dataset documenting images of cultural ornament designs. The images and their annotations are based on an open access archive of *The Grammar of Ornament* (London, 1856), an influential 19$^{th}$-century book of ornament designs by the British architect and designer Owen Jones [1]. The ornament designs originate from nineteen human cultures, including Indigenous Tribes, Greek, Egyptian, and Chinese (Figure 1). As a design-inspired dataset, JONES-19 can serve as a benchmark for various research areas, such as architectural design analysis, art-historical research, cultural heritage, visual recognition, data-efficient learning, and other. In this paper, we describe the curation of the JONES-19 dataset and report a baseline classification benchmark.

Our aim is to introduce an image dataset that poses important challenges at the intersection of ML and art and design: a small sample size of 1,901 images; a specialized class of human-designed artifacts rather than natural scenes or common objects in-context; imbalanced class distribution and high visual diversity across the nineteen cultures; fine-grained distinctions across per-class images and across classes, characterized by fine details involving line patterns, reliefs, and colors. In addition to well-established research problems – image localization, segmentation, classification, and representation learning – our dataset raises more concrete learning challenges related to design. Two such challenges are, learning the symmetry groups that characterize the ornament designs of a culture and modeling design similarities between cultures that may ground art-historical knowledge or evidence about design exchanges. Section 2 provides a background for this work and describes related studies. Section 3 describes the acquisition and curation process of the JONES-19 image dataset. Section 4 describes our baseline analysis work that focuses on the problem of cultural style classification.

## 2 BACKGROUND AND RELATED WORK

A primary motivation for this work is a long-engrained focus of the ML community on large image classification datasets that document everyday objects and natural scenes, with little coverage of design collections documenting human-designed objects that embody diverse visual-spatial languages. The JONES-19 dataset contributes towards bridging this gap, with the overarching goal of broadening the application of ML in applied arts, architecture, visual culture or humanities.

### 2.1 Design Collections as Datasets

The JONES-19 dataset is a "design collection," namely, a set of pictures or drawings that capture pattern designs of different human cultures. Historically speaking, the proliferation of design collections overlaps with a broader adoption of technologies for printing and graphic design which contributed to the public dissemination of art, design and architecture catalogues in the 18$^{th}$ and 19$^{th}$ centuries. *The Grammar of Ornament* by Owen Jones is among those catalogues that focus exclusively on ornament designs. The "ornament" – as it appears in building decoration and other functional yet beautiful artifacts – remains to this day one of the most important and extensively studied elements of composition in architecture and related pictorial arts [6, 11, 12]. In the case of architecture, the specificity of the book (i.e., that it focuses on ornament designs) is particularly important because the majority of historical collections in the field contain illustrations of plans, elevations, building components (e.g., doors, windows, columns) and structural systems. Some examples are Langley's

*The City and Country Builder's and Workman's Treasury of Designs* (1740), Durand's *Précis of the Lectures on Architecture* (1802–5), Neufert's *Architect's data* (1936), and Engel's *Structure Systems* (1967), among many others.

The *Grammar of Ornament*, however, or any other historical design collection, is not by itself a "dataset" in the canonical meaning of the term in ML and data science. Indeed, the examination of this collection leads us to a broader and more important question: What makes a design collection different from a traditional ML dataset? Design collections, including *The Grammar of Ornament* we study in this paper, exemplify the following characteristics which distinguish them from the datasets we commonly find in ML and data science research:

- **A small sample size** compared to the large-scale deep learning benchmarks that contain thousands or millions of images [22, 26–30]. This is common in historical collections that were created manually by human experts (architects, art critics, engineers, collectors), an experience acquired over many years of training and observation.
- **Imbalanced class distribution** resulting primarily from the data-scarce nature of architecture and most design fields. ML algorithms are prone to focusing on the larger classes within a dataset and are much worse at generalizing over new samples that belong to underrepresented classes.
- **Intricate visual details and pattern diversity** across cultures as well as within each culture (e.g., fine-grained distinctions based on line patterns, colors, or reliefs) which, in combination with a small sample size, poses a significant challenge to existing ML models. Relevant active research areas are fine-grained visual classification and domain adaptation for medical imaging and for images of plant specimens [28, 32].
- **Incompatible formats** for direct application of learning algorithms as any design collection must be digitally curated to become usable in ML research. In contrast, most ML benchmarks are created from existing digital images (or other data types) that are automatically crawled from the internet or extracted from public/private digital databases.

These characteristics render the JONES-19 dataset and analogous historical design collections an important and largely underrepresented source of potentially useful and impactful datasets for ML research for design and cultural applications.

### 2.2 Datasets of Ornaments

Insofar as ornaments are concerned as a distinct class of objects, the growing interest in how ML can be applied to digital culture and heritage [7, 4, 21] has resulted in several large publicly available databases of ornaments. Examples include the Y90s Database of Ornament (textual ornaments in the fin-de-siècle little magazines) [14], the Compositor database (including over a million ornaments extracted from 18th c. English books) [5, 15], the Maguelone database (18th c. ornaments printed in France and Europe) [16], and the Broadside ballads database (woodblocks extracted from English ballads) [17]. These databases originate from books that were digitized with fully automated or user-driven segmentation methods or recorded by manual methods [9, 10, 13]. The more recent Rey's Ornaments dataset [3] contains images of typographical ornamentation (i.e., woodblock ornaments) published by Marc-Michel Rey (1720-1780) in the 18th-century. Similar to our case, it is a highly imbalanced small size set of images depicting ornaments with fine design and artistic details. Unlike in our case, however, the images are low resolution, line-based, in black or greyscale colors, and do not specify information about the cultural origin of the designs they depict. There are also a number of large datasets with images of artworks, photographs, or other illustrations; for example, MAMe [8], ArtBench [18], and PHAROS [19] and the Museum of Modern Art (MoMA) digital collection [20]. In contrast, JONES-19 contains exclusively images of ornament designs, a specificity that provides concrete modeling challenges for ML research. To the best of our knowledge, there is no publicly available high-quality image dataset of ornament designs annotated based on their cultural origins.

## 3 THE JONES-19 DATASET

The JONES-19 dataset is composed of images and annotations of ornament designs taken from the original publication of *The Grammar of Ornament* (London, 1856) by Owen Jones. The digital version of this publication is provided open access by The Getty Research Institute [1]. We additionally consulted the contemporary commentaries, annotations, and curatorial decisions in the newest edition of the book published by Princeton University Press [2].

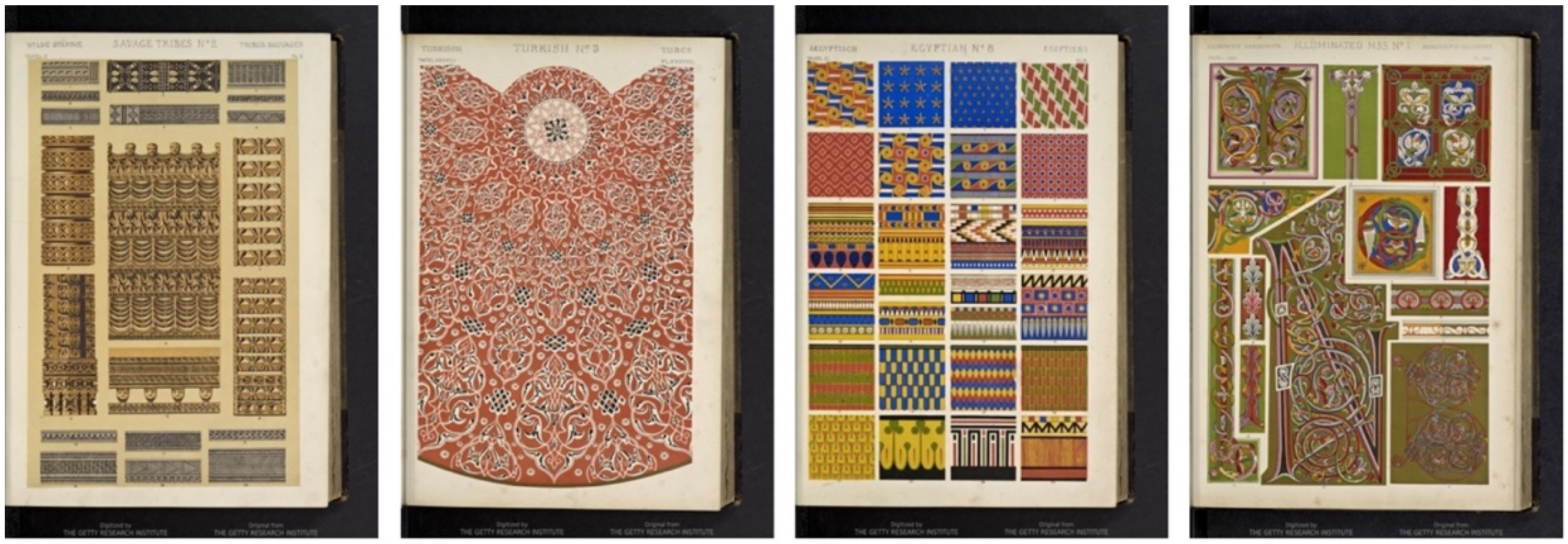

**Figure 2: Examples of scanned pages from an open-access archive for *The Grammar of Ornament* [1].**

### 3.1 Data collection and cleaning

#### *3.1.1 Pages*

The Getty Research Institute provides scanned copies of all the pages of the 1856 edition of *The Grammar of Ornament*. All pages are given on black background, in the standard format shown in Figure 2. The pages are close to 5.5K pixels on the longest dimension. The ornament designs are presented in color with varying sizes and boundary shapes. The largest ornaments cover nearly a full page, whereas the smaller ornaments can be around 280 pixels in width. Each page is annotated with a title (the culture's name), the Roman numeral of the plate, and each ornament is indexed with a number.

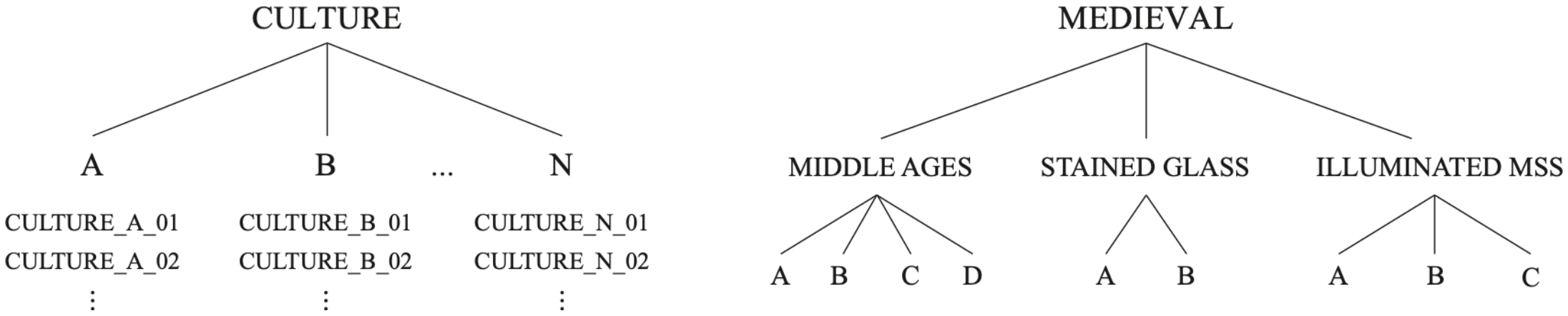

**Figure 3: (left) Structure of image groupings per culture. The lowest tree level, "CULTURE_A_01," "CULTURE_A_02," etc., are individual images. (right) "Medieval" has an additional tree level with three classes: "Middle Ages," "Stained Glass," and "Illuminated MSS"; each class is further structured into, respectively, four, two and three image groupings.**

#### *3.1.2 Image samples*

In the original book, the ornaments of all nineteen cultures are divided into two or more groups; the diagram in Figure 3(left) illustrates this grouping. The dataset preserves this grouping in the file organization and naming structure. The pages of the Medieval culture have an additional hierarchical layer illustrated in Figure 3(right). The Medieval culture is a superclass of three distinct cultures called Middle Ages, Stained Glass, and Illuminated MSS. The images in the dataset and the associated metadata files reflect this hierarchical layer but the same label Medieval is assigned to all the images to preserve the name of the superclass (this is also consistent with how the Medieval culture is indexed in the book).

### 3.1.3 *Image extraction*

The ornaments were extracted from the pages of the book by a combination of automated and manual methods. Figure 4 describes the extraction process in a diagrammatic way. We used the Segment Anything [24] automated image segmentation system. The model effectively outputs image masks for irregular and inconsistent shapes and sizes, though it requires a long processing time due to the model's complexity. The segmentation process consists of four tasks: (a) downsampling images for faster processing, (b) generating and filtering segmentation masks based on area constraints, (c) scaling masks back to the original image size, and (d) cropping the images. For pages with consistent grids, our process selectively employs image processing methods in OpenCV [25] for their speed and simplicity. After automation, we manually reviewed and curated the resulting image files. Images that were missed or poorly segmented, especially those with complex boundaries, were individually extracted using the Preview application on MacOS. Finally, a peer review verified the resulting dataset's completeness by comparing the segmented images to the original pages of the book.

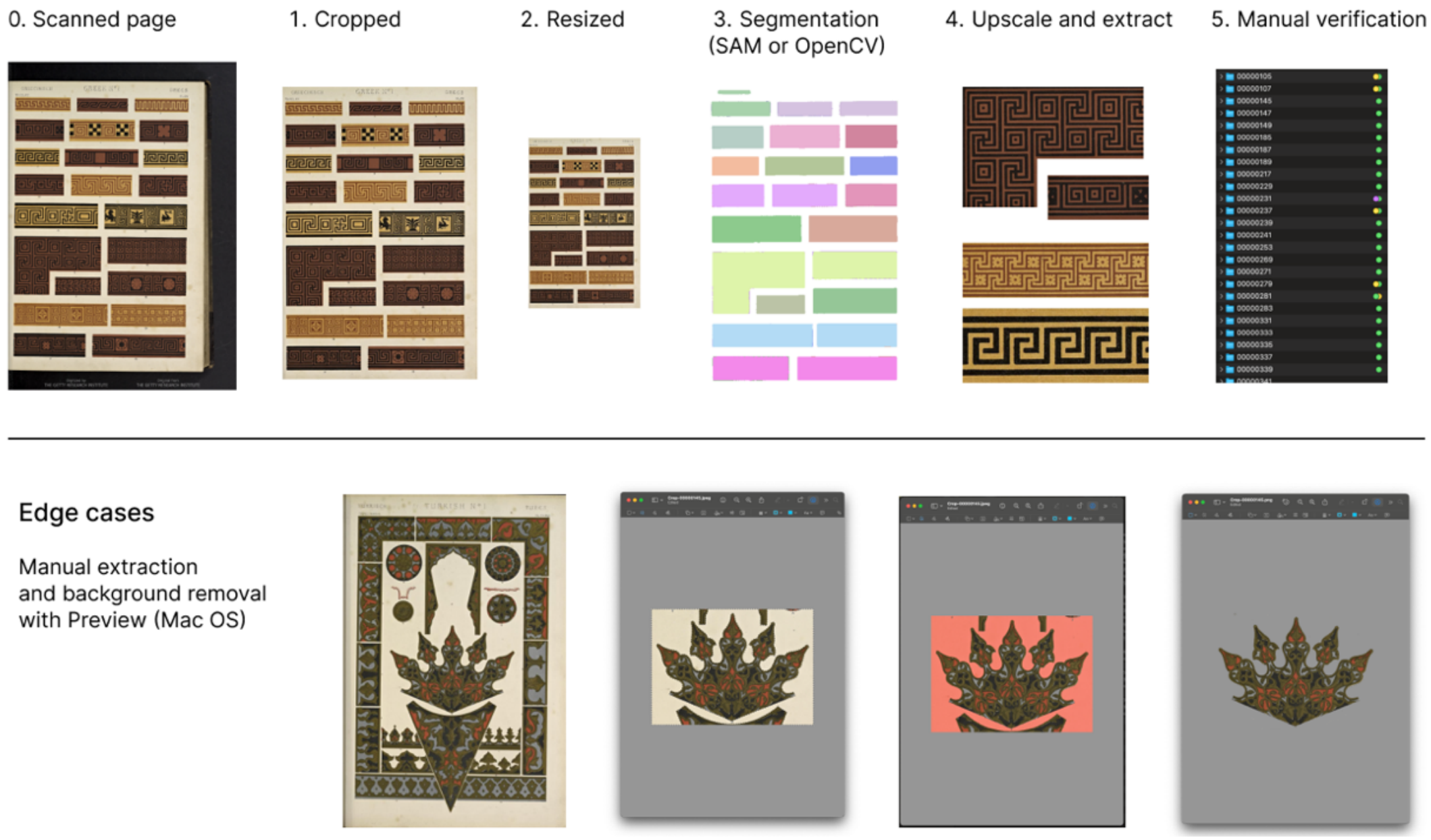

**Figure 4: Process of extracting images from a scanned book page.**

### 3.1.4 *Image filtering decisions*

The JONES-19 dataset contains nearly all images of ornament designs of the original book. The dataset excludes ornament designs that are not depicted in a flat orthogonal projection; for example, see the Egyptian columns on Plate VI. It also excludes ornaments that are not presented as standalone designs but are printed on objects, such as weapons and crowns. It also excludes page-size ornaments such as those on Plate XX consisting of individual labelled ornaments for which we could note discern a fixed boundary. Similarly, it excludes page-size ornaments (e.g., Plates XXV and LXXVII) for which we could not determine from the book whether to treat them as single large ornaments or sets of individual ornaments (in contrast, single page-size ornaments on Plates XXXVIII and LXVI are included). The dataset additionally excludes ornaments whose boundary lines we could not clearly determine. This typically occurs with ornaments featuring decorative rather than clear continuous boundaries (e.g., Turkish ornament 3 on Plate XXXVII, and ornaments on Plates LXXII, LXXIII, and XC). Our dataset also excludes chapter XX called "Leaves and Flowers from Nature," which contains drawings of natural leaves and flowers rather than artificial ornament designs. Last, the original book contains composite ornaments, i.e., two or more distinct designs drawn next to or on top of each other, that Owen Jones treats as one – for example, see the Indian Plate LV. In JONES-19, we have decomposed composite ornaments whose boundary lines can be clearly determined. This increases the total number of image samples for certain cultures, such as Indian and Chinese.

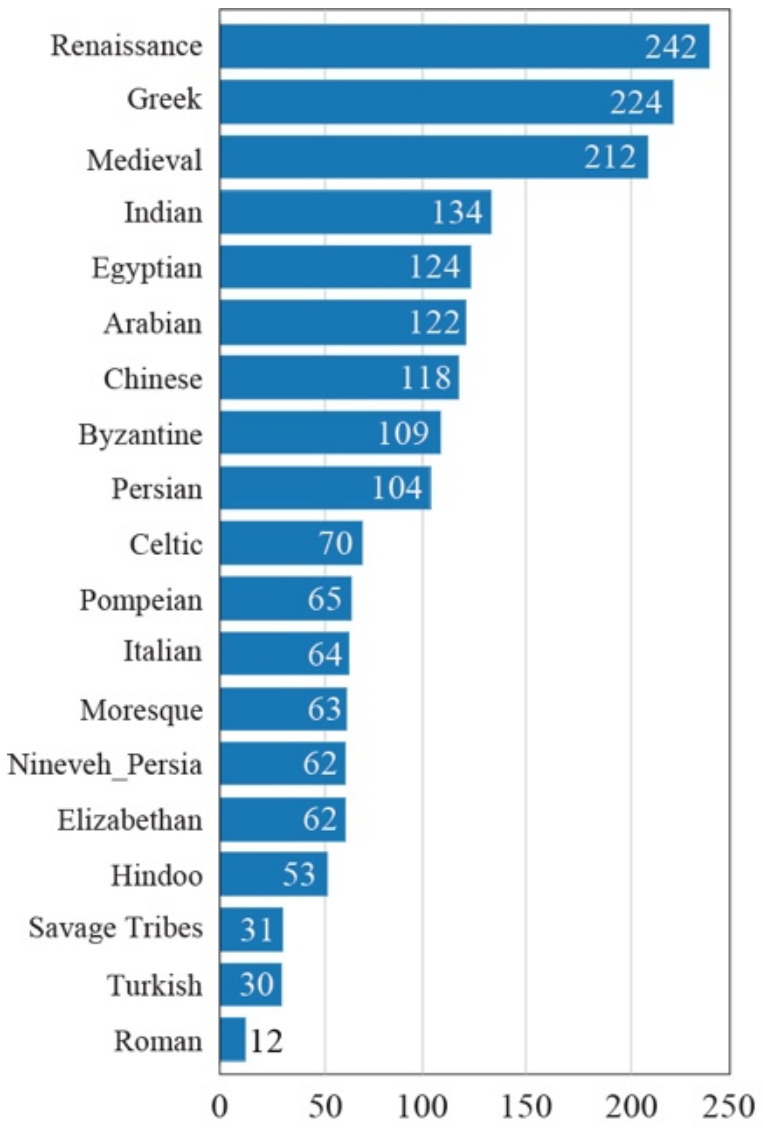

(a)

| JONES-19 Size | |
|---|---|
| #of Ornaments in GrO | 2,071 |
| #of Ornaments Excluded | 170 |
| Total | 1,901 |

(b)

| JONES-19 Composition | |
|---|---|
| #of Classes | 19 |
| Source of GT | GrO-1856 |
| #of Images/Class | 12 – 242 |
| Hierarchy Level | 1st |

(c)

**Figure 5: Dataset statistics. (a) Per class distribution of image samples shows a class-unbalanced dataset. (b) JONES-19 consists of 1901 samples out of the 2071 images in the original book (here, GrO). (c) There are no hierarchical levels in the dataset.**

### 3.2 Dataset characteristics

Details about the JONES-19 dataset statistics are shown in Figure 5. The full dataset consists of 1,901 images of distinct ornament designs. The dataset is not a comprehensive and accurate representation of ornament designs throughout the known world; it covers the ornament designs selected by the author in the 19th century. We have not performed any experiments to test whether the classification models generalize to ornament designs outside of the primary cultures represented in the book. We believe that the existing image dataset with its nineteen cultures already poses interesting and significant challenges for ML research in architecture, art or design. There are 2,071 images of ornament designs in the original book. We excluded 170 images following the filtering principles outlined in Section 3.1.3. Excluding chapter XX (see above), the number of ornaments that are not in JONES-19 represent about 8% of the total number of images in the book. Figure 5(a) shows the class-unbalanced distribution of the dataset. For example, just three cultures, namely, Renaissance, Greek, and Medieval, are the largest classes in JONES-19, representing about 35% of the total dataset size.

All image samples are labelled with a single label of the cultural style they belong to, casting the benchmark analysis described in Section 4 as a single-label classification task asking for the primary cultural style.

## 4 CLASSIFICATION BASELINE AND ANALYSIS

In this section, we evaluate the suitability of the JONES-19 dataset for training classifiers by analyzing baseline performance. Our goal is to assess the dataset's complexity through classification accuracy and to gain insight into inter-class relationships–particularly cultural similarities–by examining patterns in misclassification errors. We report details of our results in Figure 6 and in the Appendices A.1–A.3.

### 4.1 Model Setup

We benchmark the performance of ResNet18 and ResNet50 [22] for transfer learning in a classification task using two input image resolutions, 224 x 224 and 448 x 448 pixels. We use models pretrained on ImageNet-1k [23], replacing their final classification layers with randomly initialized heads for the nineteen classes in our dataset. Each configuration is fine-tuned using 10-fold cross-validation (one fold per run) with a fixed random seed. For each fold, the model is trained for up to 40 epochs using Adam optimizer [31]. Early stopping is applied based on validation loss to prevent overfitting. Details of hyper-parameters can be found in the Appendix A.2.

The images are resized, cropped, and normalized to match the ImageNet mean and standard deviation. For data augmentation, we apply a random rotation of up to ±5 degrees and horizontal flip in 50% of the cases. Training images are randomly cropped, while test images are center-cropped. Our dataset contains images significantly larger than the input resolutions, allowing each image to yield multiple unique training examples per ornament through cropping. Given an average image size of 680×555, we estimate the dataset size could effectively increase by a factor of ~7. This estimate is based on the approximate number of crops that can be extracted from each image. While we did not implement this augmentation method in this paper, it presents a promising direction for future work.

Given the high class-imbalance of the dataset, we adopt a stratified sampling approach. Specifically, for each fold we first perform an outer stratified split using `StratifiedKFold` to divide the data into training and test subsets, ensuring each fold preserves class distribution. Then, within each training set, we apply a `StratifiedShuffleSplit` to create an 80/20 training/validation split, again reflecting the class distribution of the entire dataset, maintaining balanced class representation. Each fold included distinct subsets for training, validation, and testing, with all samples appearing exactly once in the test dataset across folds. In particular, this two-stage stratified sampling approach ensures that all nineteen classes are present at roughly equal proportions in every training, validation, and test subset, at roughly similar proportions. As is common with class-imbalanced datasets, we use a class-weighted loss. For this, we calculate class weights as the inverse of class frequencies and apply these weights during training using a weighted cross-entropy loss criterion, ensuring underrepresented classes have proportionally greater influence on training. We utilize 10-fold cross-validation for data splitting to provide a fair estimate of model performance, stratifying the data as described above.

Validation loss and accuracy are monitored during training to select the best-performing model checkpoint for each run. To account for class imbalance, we evaluate the classification performance in terms of macro-average recall, which averages per-class recall and mitigates the effect of class imbalance on performance reporting. For each configuration, we repeat each experiment ten times and report the mean balanced accuracy and standard deviation.

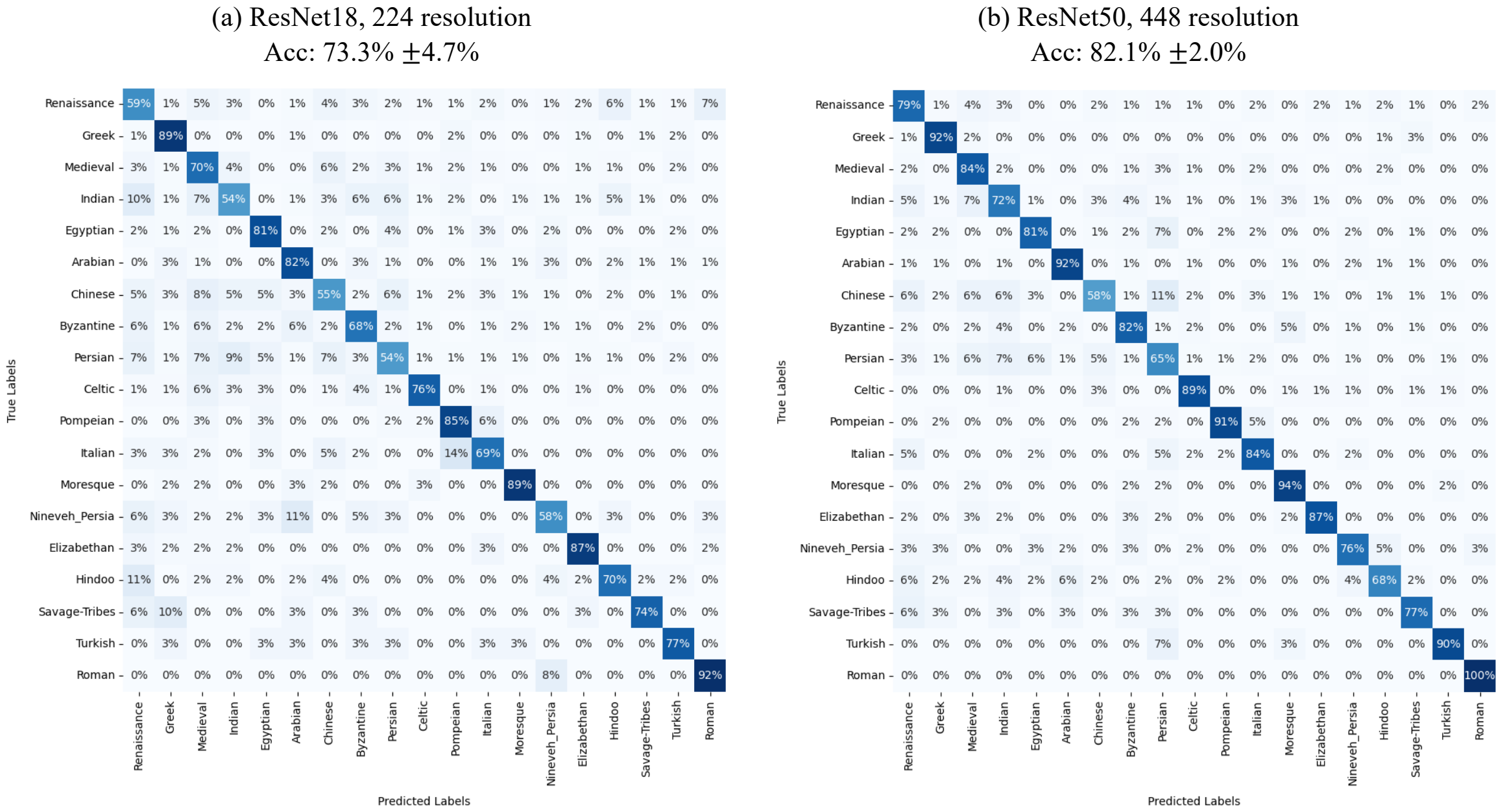

**Figure 6: Confusion matrices for the worst and best performing baseline classifiers on JONES-19. Subfigure captions specify the input image resolution and model name. The dimensions of the matrices are sorted based on class count. Accuracy is measured as mean balanced accuracy over ten runs.**

## 4.2 Analysis and Results

### *4.2.1 Model classification accuracy*

The confusion matrices of aggregated balanced accuracies across all experiments are in Appendix A.1. Two of these matrices–the worst and best model performances–are also printed in Figure 6. All models achieve accuracies between 73% and 82%, a significant improvement over random guessing, which would yield approximately 15% accuracy for nineteen classes. The results show that both increasing input resolution and using a larger network improves classification accuracy. ResNet50, in particular, outperforms ResNet18 with greater improvements in underrepresented classes. ResNet50 achieves 81.1% (±2.2%) with 224×224 resolution and 82.1% (±2.0%) with 448×448 resolution, compared to ResNet18, which achieves 73.3% (±4.7%) and 76.7% (±5.4%) for the same resolutions. Using a shallower network and lower resolution often leads to confusions between culturally similar classes. For example, ResNet18 with 224×224 resolution achieves 69% accuracy for Italian ornaments, with 16% of misclassifications attributed to Pompeian. This issue is mitigated with ResNet50 at 448×448 resolution, which improves the accuracy for Italian ornaments to 85%, reducing misclassifications to 2% for Pompeian and 5% for Renaissance. These improvements demonstrate the ability of higher resolutions and deeper architectures to capture intricate details, such as line thickness, reliefs, and complex symmetry.

### *4.2.2 Inter-class relationships and cultural similarity*

An important challenge emerging from this investigation of the JONES-19 dataset is determining whether the models' misclassifications result from cross-cultural similarities between ornament designs. Owen Jones devotes considerable attention throughout the original book in explaining design relationships between the ornaments of different cultures. For example, many Pompeian ornament designs have a Greek character [2: 118] while others remind us of the variety of Byzantine, Arabian and Moresque mosaics [2: 119]; features of Celtic ornaments (e.g., “Z” patterns, angulated step-like lines) appear in both the Chinese and Illuminated MSS ornament [2: 295]; and so forth. While performing an extensive systematic study of how our baseline models’ misclassifications may be explained in terms of design features is outside the scope of this paper, here we highlight a few of the largest and most evident misclassifications in our matrices. Along with our dataset submission, we also provide a complementary folder of cultural misclassifications for future research.

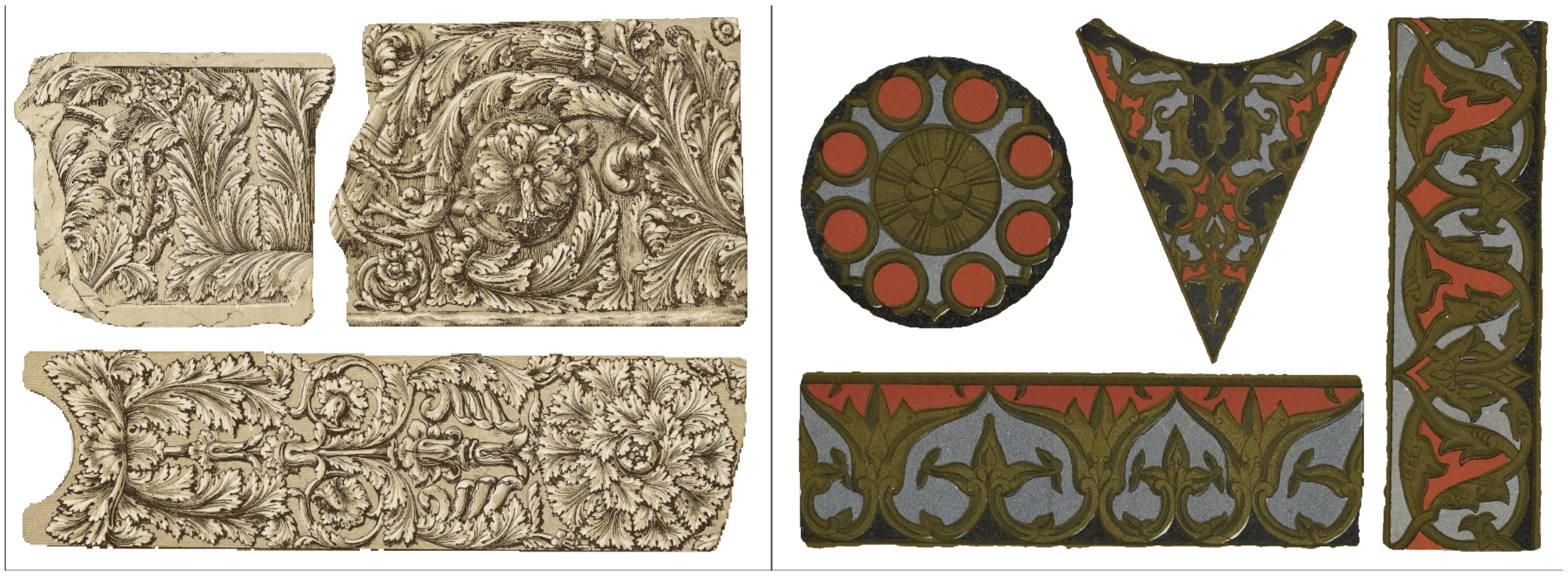

**Figure 7: Image samples from the Roman (left) and Turkish (right) ornament classes.**

In particular, the classification accuracies for the Persian and Chinese cultures are relatively low–65% and 58%, respectively–and, with our best performing model, a Chinese ornament may be classified as Persian 11% of the time. The Persian and Chinese ornaments share design characteristics such as floral elements and colorful geometric tilings. Similarly, a subset of the Indian ornament is consistently misclassified as Medieval or Renaissance across all models, two of the largest classes in the dataset. Examples of Chinese ornaments misclassified as Persian and Indian ornaments misclassified as Medieval are in the table in Appendix A.3. Owen Jones does not explicitly describe any design similarities between these cultures (instead, he describes relations between the Indian and the Arabian and Persian ornaments). This implies that misclassifications by deep learning models may or may not be causally derived from provable or historically documented exchanges between cultures. Owen Jones describes design exchanges between Byzantine, Medieval, and Renaissance styles, which are also misclassified by our models, as reflecting their shared stylistic roots. This underscores the difficulty of learning to discern subtle design variations between neighboring cultural traditions, often resulting from exchanges on themes around religion, mythology, nature or geometry. Small, underrepresented classes such as Roman (100%) and Turkish (93%) are classified with very high accuracy in all experiments; Figure 7 shows image samples from both of these cultures. This likely reflects their distinct motifs and lower in-class visual diversity making it easier for the models to learn to distinguish amongst their image samples; for example, see the “stone cuts” and reliefs embedded in the Roman ornament in Figure 7(left).

## 5 CONCLUSION

As a high-quality image dataset, JONES-19 poses numerous challenges as a benchmark for computer vision classification tasks and for research at the intersection of ML and architecture, art and design. Of significant importance are the following dataset characteristics: small sample size, high intra-class visual diversity, imbalanced class distribution, and fine-grained image distinctions based on line patterns, reliefs, and colors. Because JONES-19 is a design-inspired dataset that documents a specialized type of design object (i.e., ornament designs), it differs significantly from large datasets that document common objects in-context or natural scenes. It also differs from large galleries of artworks from different periods, styles or artists that we typically find in the digitized archives of museums and libraries. Thus, the specificity that characterizes JONES-19 can serve as a new benchmark for various research fields, such as fine-grained visual recognition, data-efficient learning, domain adaptation, art-historical research, architectural style analysis, and cultural heritage. The analysis in Section 4 additionally underscores important future directions for the problem of cultural ornament classification and the analysis of cross-cultural influences in ornament design. Specifically, JONES-19 can serve as a platform for investigating cross-cultural design exchanges computationally, to delve deeper into the intelligence of discerning one design from another, a human skill that remains central in architecture, art or design.

## ACKNOWLEDGMENTS

The contributions of the authors in this publication are as follows: Conceptualization: A.H.; Methodology: A.H., L.P., D.A-M.; Software: A.H., L.P.; Validation: A.H., D.A-M., S.S.; Formal analysis: A.H., L.P., D.A-M.; Investigation and Data curation: L.P.; Writing – original draft: A.H.; Writing – review & editing: L.P., D.A-M., S.S.; Supervision and Project administration: A.H.; Funding acquisition: A.H., D.A-M.

We thank Peggy Yin (Harvard College) and Corey Wang (Harvard SEAS) for their technical assistance. Funding for this paper was provided by the Harvard Data Science Initiative.

# A APPENDICES

## A.1 Confusion Matrices

(a) ResNet18, 224 resolution
Acc: 73.3% ±4.7%

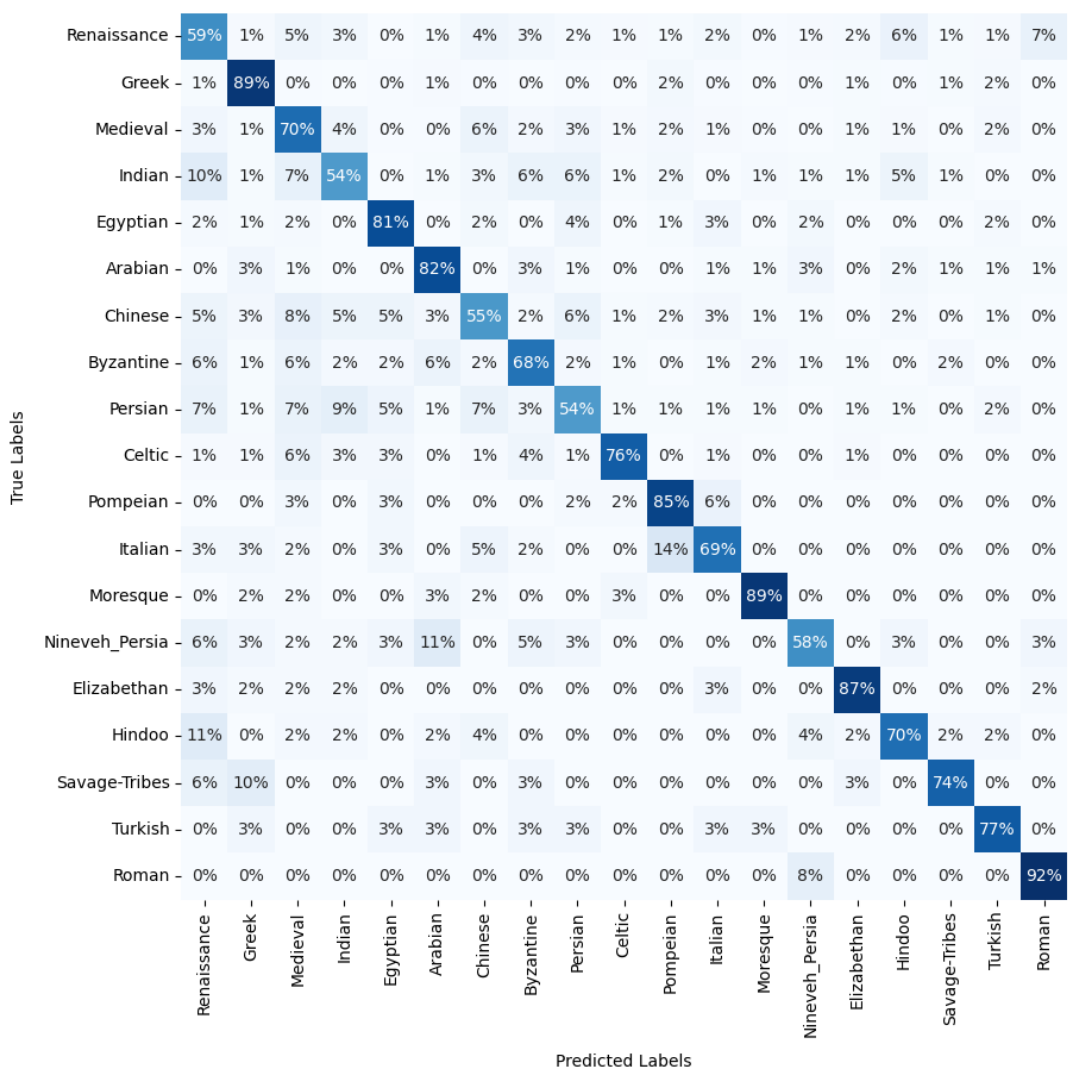

(b) ResNet18, 448 resolution
Acc: 76.7% ±5.4%

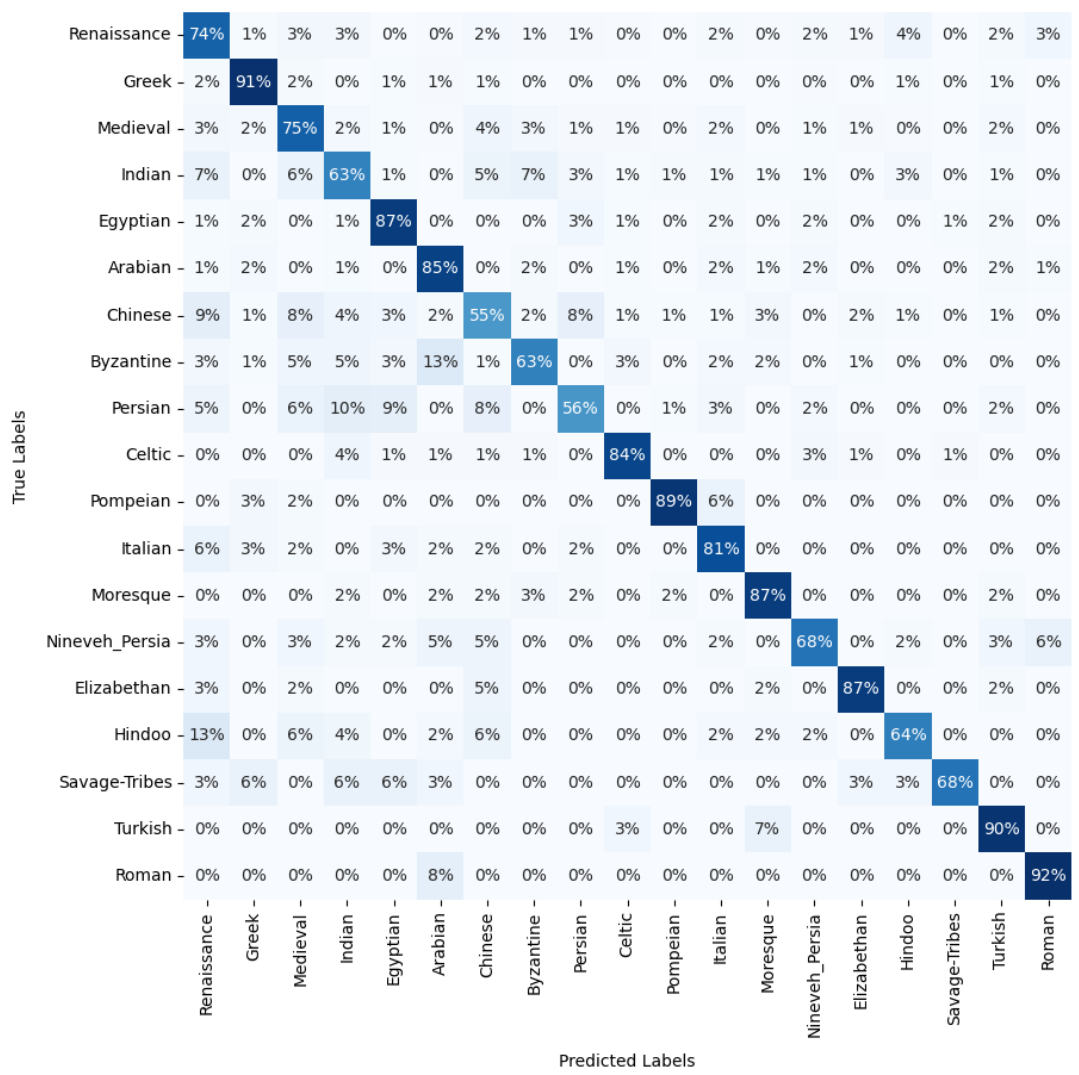

(c) ResNet50, 224 resolution
Acc: 81.1% ±2.2%

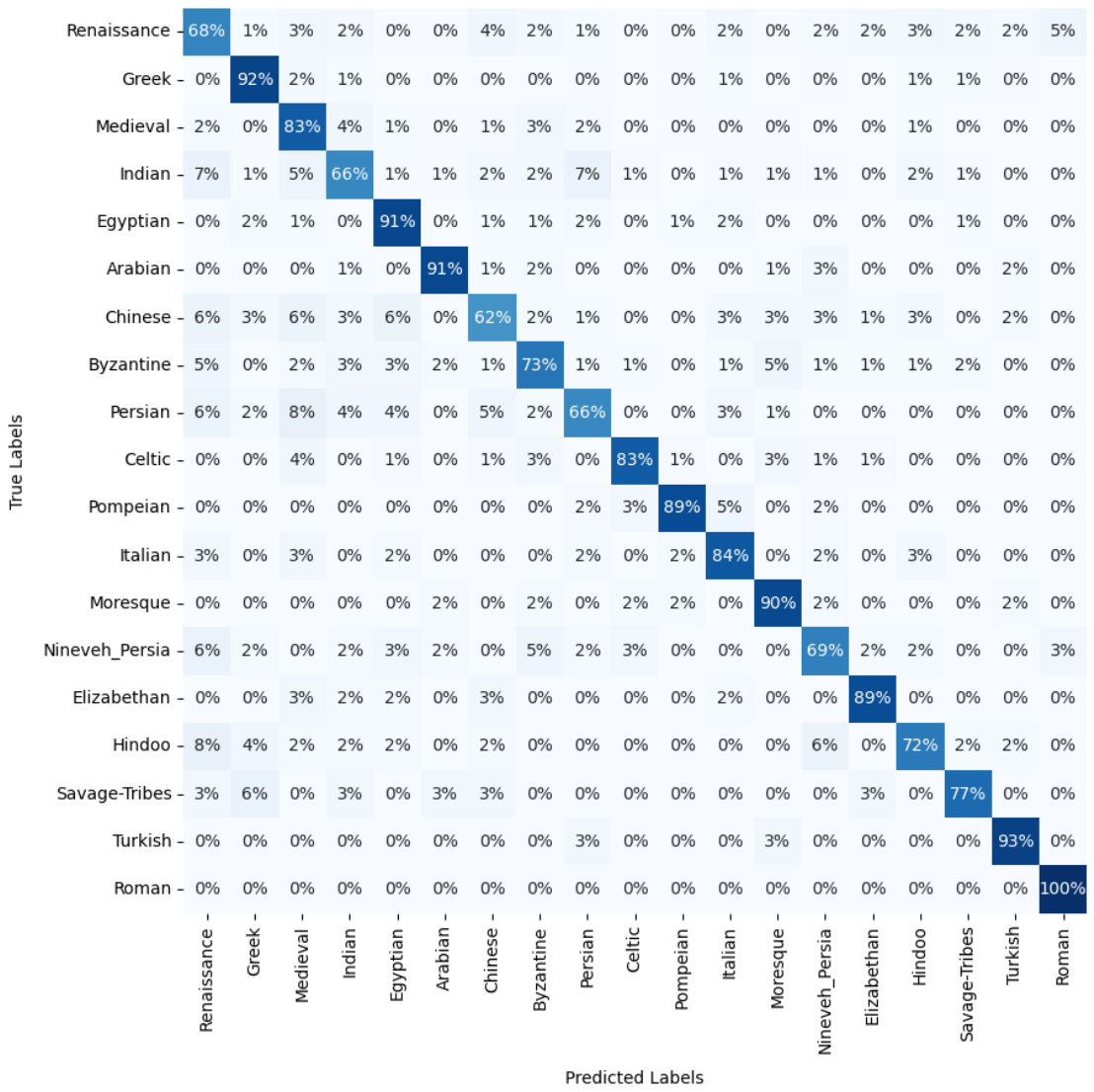

(d) ResNet50, 448 resolution
Acc: 82.1% ±2.0%

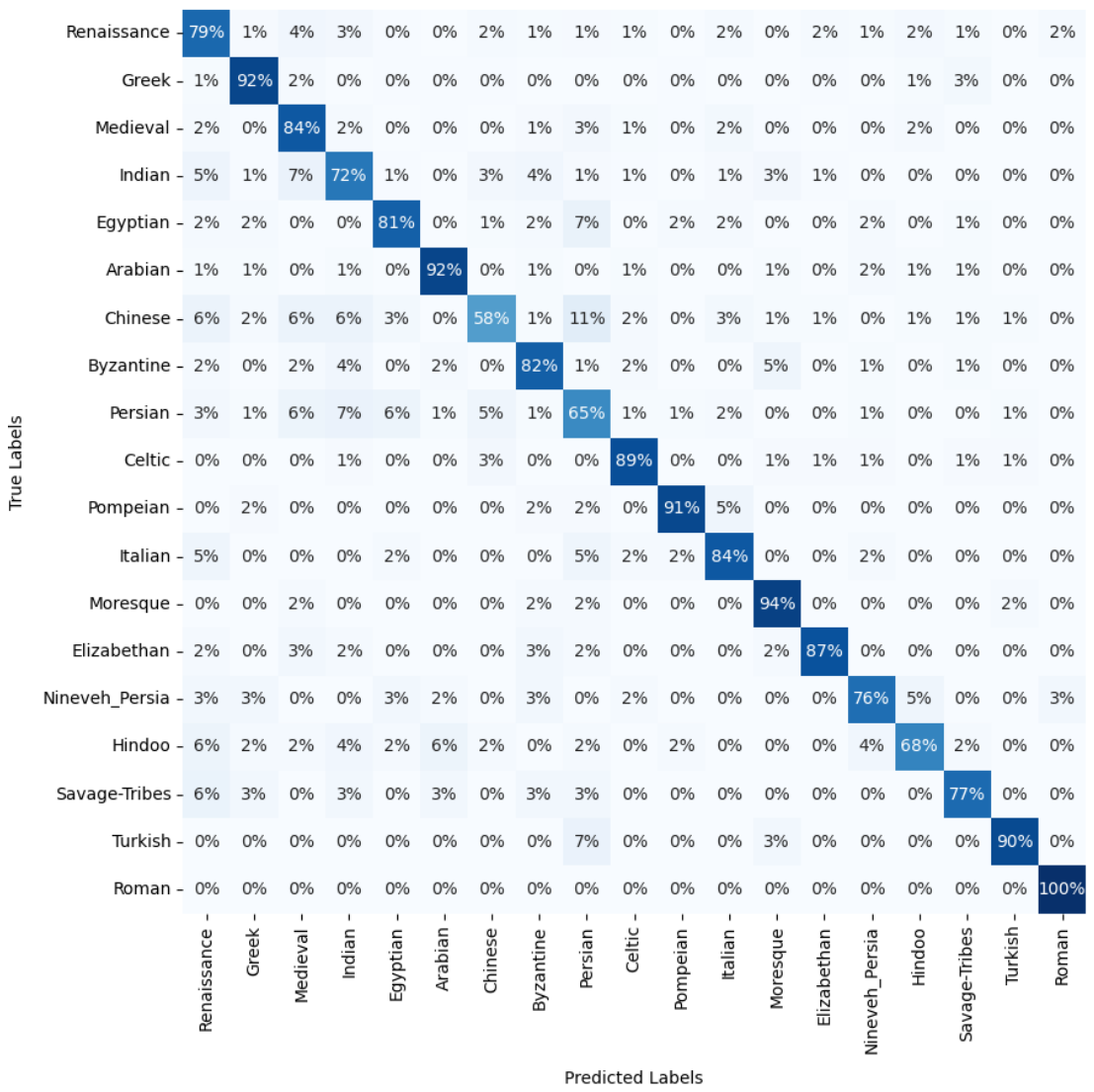

**Figure 8: Confusion matrices for two variants of each of our baseline classifiers on JONES-19. Subfigure captions specify the input image resolution and model name. The dimensions of the matrices are sorted based on class count. Accuracy is measured as mean balanced accuracy over ten runs.**

### A.2 Hyper-Parameters table

Table 1: Hyper-parameters used for training the four baseline models on JONES-19.

| Models | Pre-training | Balancing | Resolution | Epochs | Folds | Batch Size | Learning Rate | Optimizer | Loss |
|---|---|---|---|---|---|---|---|---|---|
| ResNet18 | ImageNet-1k | Stratified K-Fold + Stratified Shuffle Split | 224 | $\leq 40$ | 10 | 32 | $1\times10^{-3}$ | Adam | Weighted Cross Entropy* |
| ResNet18 | ImageNet-1k | Stratified K-Fold + Stratified Shuffle Split | 448 | $\leq 40$ | 10 | 32 | $1\times10^{-3}$ | Adam | Weighted Cross Entropy* |
| ResNet50 | ImageNet-1k | Stratified K-Fold + Stratified Shuffle Split | 224 | $\leq 40$ | 10 | 32 | $1\times10^{-3}$ | Adam | Weighted Cross Entropy* |
| ResNet50 | ImageNet-1k | Stratified K-Fold + Stratified Shuffle Split | 448 | $\leq 40$ | 10 | 32 | $1\times10^{-3}$ | Adam | Weighted Cross Entropy* |

* Class weights were set as the inverse of class frequencies to address label imbalance during training.

### A.3 Misclassifications table

Table 2: Examples of misclassifications produced by ResNet50 (448).

| Culture A | Culture B | Culture A misclassified as Culture B* |
|---|---|---|
| Chinese | Persian | 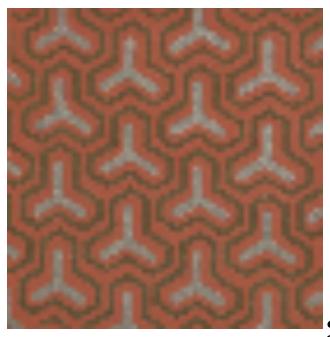, 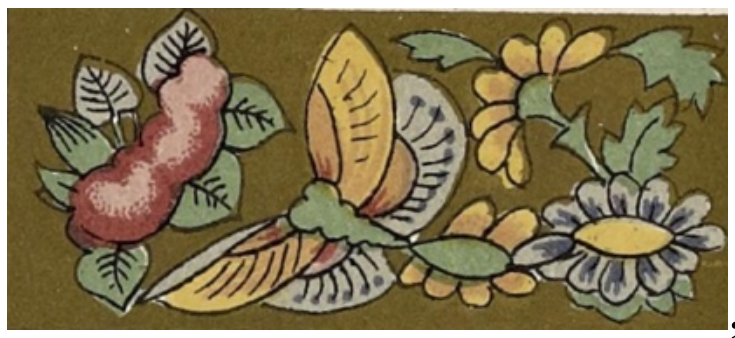, 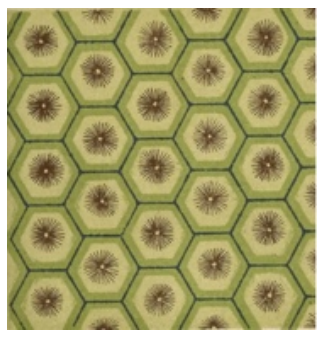 |
| Indian | Medieval | 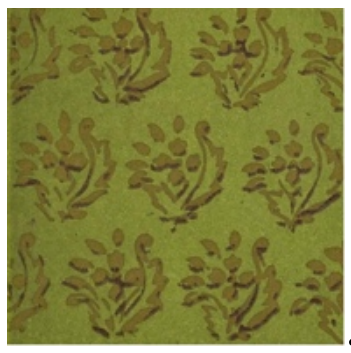, 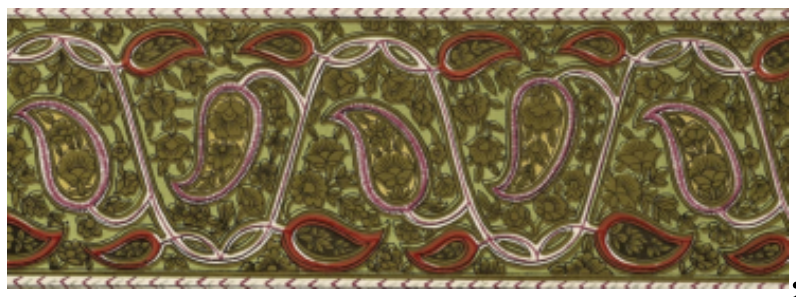, 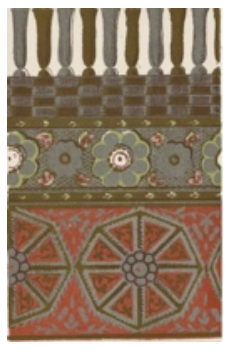 |
| Greek | Savage Tribes | 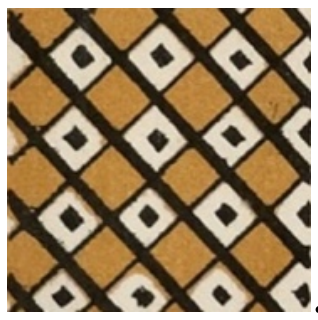, 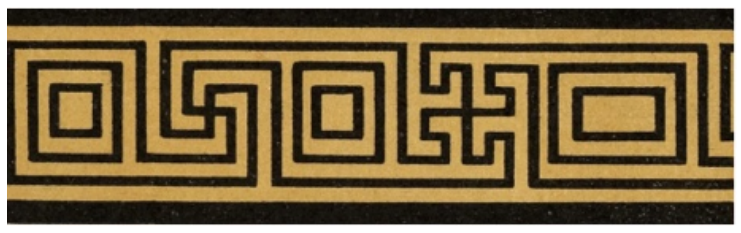, 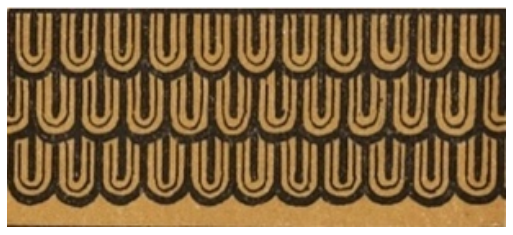 |

* e.g., Chinese ornament misclassified as Persian ornament.